\pdfoutput=1
\documentclass[conference, compsoc]{IEEEtran}
\IEEEoverridecommandlockouts
\makeatletter
\newcommand*\titleheader[1]{\gdef\@titleheader{#1}}
\AtBeginDocument{%
  \let\st@red@title\@title
  \def\@title{%
    \bgroup\normalfont\normalsize\centering\@titleheader\par\egroup
    \vskip1ex\st@red@title}
}

\renewcommand\footnoterule{%
  \kern-3\p@
  \hrule\@width.4\columnwidth
  \kern2.6\p@}

\makeatother

\usepackage{graphicx,array}
\usepackage{amsmath,amssymb,amsfonts}
\usepackage{multirow}
\usepackage[table]{xcolor}
\usepackage[inline]{enumitem}

\title{FaceLiVT: Face Recognition using Linear Vision Transformer with Structural Reparameterization For Mobile Device}
\usepackage{hyperref}
\hypersetup{
    colorlinks=true,
    linkcolor=blue,
    filecolor=magenta,      
    urlcolor=cyan,
    pdftitle={Overleaf Example},
    pdfpagemode=FullScreen,
    }

\newcommand\blfootnote[1]{%
  \begingroup
  \renewcommand\thefootnote{}\footnote{#1}%
  \addtocounter{footnote}{-1}%
  \endgroup
}                                                                                       

\author{\IEEEauthorblockN{Novendra Setyawan\textsuperscript{1,3}, Chi-Chia Sun\textsuperscript{*,2}, Mao-Hsiu Hsu\textsuperscript{1}, Wen-Kai Kuo\textsuperscript{1}, Jun-Wei Hsieh\textsuperscript{4}}
\IEEEauthorblockA{\textit{\textsuperscript{1}Department of Electro-Optics Engineering, National Formosa University, Taiwan}\\
\textit{\textsuperscript{2}Department of Electrical Engineering, National Taipei University, Taiwan} \\
\textit{\textsuperscript{3}Department of Electrical Engineering, University of Muhammadiyah Malang, Indonesia} \\
\textit{\textsuperscript{4}College of Artificial Intelligence and Green Energy, National Yang Ming Chiao Tung University, Taiwan} 
}
}

\titleheader{\vspace{-20pt}To appear at the 2025 IEEE International Conference on Image Processing (ICIP), Sep 14-17 2025, Alaska, USA.}

\begin{document}
%
\maketitle
\blfootnote{Corresponding Author (email: chichiasun@gm.ntpu.edu.tw\textsuperscript{*}) \\ This work was supported by the National Science and Technology Council, Taiwan under Grant NSTC-113-2221-E-305-018-MY3.}
\begin{abstract}
This paper introduces FaceLiVT, a lightweight yet powerful face recognition model that integrates a hybrid Convolution Neural Network (CNN)-Transformer architecture with an innovative and lightweight Multi-Head Linear Attention (MHLA) mechanism. By combining MHLA alongside a reparameterized token mixer, FaceLiVT effectively reduces computational complexity while preserving competitive accuracy. Extensive evaluations on challenging benchmarks—including LFW, CFP-FP, AgeDB-30, IJB-B, and IJB-C—highlight its superior performance compared to state-of-the-art lightweight models. MHLA notably improves inference speed, allowing FaceLiVT to deliver high accuracy with lower latency on mobile devices. Specifically, FaceLiVT is 8.6$\times$ faster than EdgeFace, a recent hybrid CNN-Transformer model optimized for edge devices, and 21.2$\times$ faster than a pure ViT-Based model. With its balanced design, FaceLiVT offers an efficient and practical solution for real-time face recognition on resource-constrained platforms.
\end{abstract}
\begin{IEEEkeywords}
Face Recognition, Vision Transformer, Multi-Head Linear Attention (MHLA), Structural Reparameterization, Lightweight Model
\end{IEEEkeywords}
\section{Introduction}
\label{sec:intro}
Face recognition is a key technology for identity verification, widely adopted in mobile and embedded systems for applications such as device unlocking, app access, and mobile payments. In scenarios like smartphone authentication, face verification must often be performed locally \cite{chen2018mobilefacenets}. Therefore, it is critical that face verification models for mobile devices are both accurate and computationally efficient.

Although deep neural networks have significantly improved face recognition performance, their large number of parameters demands substantial memory and processing power, making deployment on resource-limited platforms challenging \cite{setyawan2024fpga}. To address this, researchers have focused on developing lightweight neural networks that strike a better balance between accuracy and computational efficiency. Lightweight CNN architectures, such as MobileNetV1 \cite{chen2018mobilefacenets}, MobileNetV2 \cite{martinez2021benchmarking}, and ShuffleNet \cite{martindez2019shufflefacenet}, have been proposed for facial verification, reducing model size and complexity. However, these models often suffer from limited receptive fields and insufficient modeling of long-range dependencies, which can lead to reduced accuracy.

Transformer-based vision models have recently achieved impressive results across many computer vision tasks \cite{dosovitskiy2020image, xie2021segformer, setyawan2025microvit, carion2020end}, mainly due to their ability to capture long-range dependencies via global receptive fields. Yet, their quadratic computational complexity makes them unsuitable for mobile applications. For instance, Vision Transformer (ViT) variants require up to 632 million parameters and 25.4 GFLOPs, making them impractical for edge deployment \cite{an2022killing, dan2023transface}. Several hybrid models, like EdgeFace  \cite{george2024edgeface}, attempt to reduce this overhead using low-rank approximations with sequential linear layers. While this reduces parameter count and FLOPs, it may also slow down inference due to extra operations.

In this paper, we propose FaceLiVT (Linear Vision Transformer for Face Recognition), a hybrid CNN–Transformer model equipped with Multi-Head Linear Attention (MHLA) and a reparameterized token mixer. FaceLiVT introduces two key token mixers: RepMix, which uses depthwise convolution with reparameterization in early stages, and MHLA, which replaces Multi-Head Self-Attention (MHSA) in later stages to reduce computational cost. Through comprehensive experiments on benchmarks such as LFW, CFP-FP, AgeDB-30, IJB-B, and IJB-C, we demonstrate that FaceLiVT achieves a strong balance between accuracy and latency, making it well-suited for mobile deployment. The key contributions of this paper are:
\begin{figure*}
    \centering
    \includegraphics[width=15.2cm]{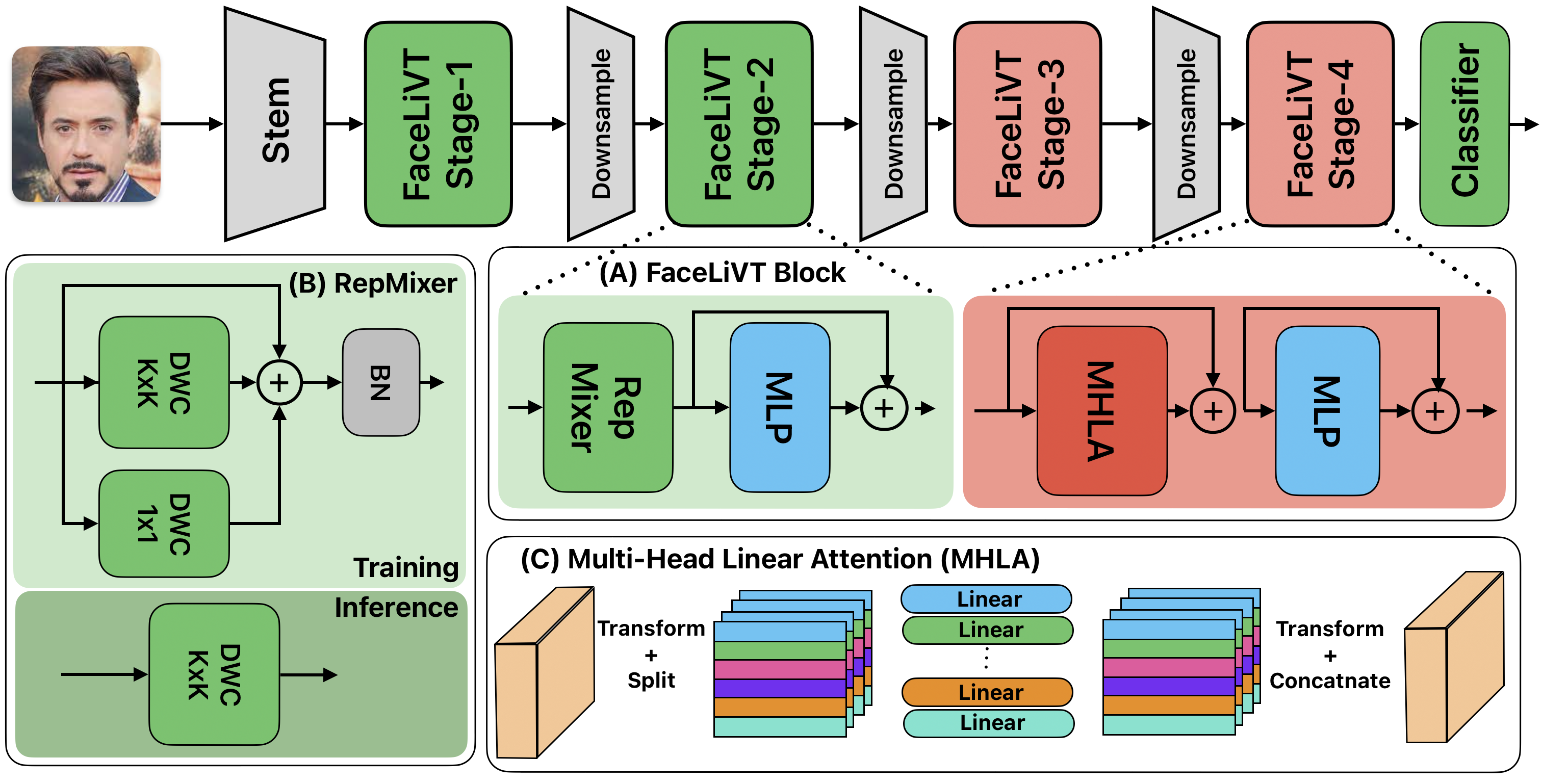}
    \caption{FaceLiVT architecture with Multi-Head Linear Attention (MHLA) and structural reparameterization. Stages 1 and 2 use the RepMix and the last stage used MHLA as token mixer. (a) FaceLiVT Block. (b) RepMix. (c) MHLA. }
    \label{fig:FaceLiVT}
\end{figure*} 
\begin{enumerate}[label=\arabic*)]
\item We propose FaceLiVT, an efficient and lightweight face recognition network that combines CNN and ViT features through reparameterization, enabling real-time performance on resource-constrained platforms.
\item We introduce a Multi-Head Linear Attention (MHLA) module to reduce computational cost with linear layers while maintaining performance. MHLA replaces Self-Attention to capture spatial correlations with low complexity. To our knowledge, this is the first Hybrid CNN-Transformer incorporating MHLA for efficient face recognition.
\item We conduct extensive experiments on challenging face recognition datasets, demonstrating FaceLiVT's superior performance over existing lightweight models, along with its efficiency in mobile inference.
\end{enumerate}

This paper is organized as follows: Section 2 presents a concise review of related works and Section 3 offers an in-depth description of the proposed FaceLiVT model. Section 4 outlines the experimental design. Section 5 concludes the paper and proposes directions for future research.

\section{Related Works}
\label{sec:relatedworks}

MobileFaceNets \cite{chen2018mobilefacenets, martinez2021benchmarking} represent a family of efficient CNN models based on the MobileNetV1 and MobileNetV2 framework, designed specifically for applications in real-time face verification \cite{martinez2021benchmarking}. MobileFaceNetV1, based on the MobileNetV1 model, has achieved an accuracy of 99.4\% on the LFW dataset, while MobileFaceNet, based on the MobileNetV2 architecture, achieved an accuracy of 99.7\% on the LFW dataset,  while maintaining a parameter count below 1 million. Inspired by ShuffleNetV2, a series of lightweight FR models termed ShuffleFaceNet was introduced in \cite{martindez2019shufflefacenet}. These models feature parameters ranging from 0.5 million to 4.5 million and have shown verification accuracies surpassing 99.20\% on the LFW dataset. 

Another approach aims to develop a face recognition model based on the original Vision Transformer (ViT) \cite{an2022killing, dan2023transface}. This ViT achieves high accuracy on several benchmark datasets but exhibits a significant complexity of 1.5 GFLOPs, rendering it impractical for mobile applications. Based on the EdgeNeXt architecture, which serves as a hybrid model integrating the strengths of Transformers and CNNs, EdgeFace \cite{george2024edgeface} is designed to decrease both the parameter count and floating-point operations (FLOPs) of this architecture, reducing parameters from 2.24 to 1.77 million and FLOPs from 196.9 to 153.9 million. This can be accomplished through low-rank linear approximation using a sequence of two linear layers, though it may result in reduced latency on mobile devices due to the use of two linear layers instead of one.
\section{Proposed Method}
\label{sec:method}
We propose FaceLiVT, a hybrid CNN–Transformer model optimized for efficient face recognition on mobile devices. It combines RepMix, a reparameterized convolutional block for local feature extraction, with Multi-Head Linear Attention (MHLA) to reduce the complexity of traditional self-attention. This design maintains high accuracy while significantly lowering inference latency and computational cost, making FaceLiVT ideal for real-time deployment on resource-constrained platforms.

\subsection{Architecture}
The overall architecture of FaceLiVT follows a macro design inspired by MetaFormer \cite{yu2022metaformer, yu2023metaformer}, consisting of two stacked residual blocks, as depicted in Fig. \ref{fig:FaceLiVT}. It begins with a stem module composed of two 3×3 convolutions with a stride of 2. Each stage contains a token mixer for spatial feature extraction and a channel mixer for refining information. Each block consists of a normalization layer and either residual or skip connections to stabilize the loss and enhance the training process. Let $X_i, X'_i$, and $X''_i \in \mathbb{R}^{H_i\times W_i\times C_i}$ represent the feature maps at stage $i$ with a resolution of $H_i \times W_i$ and $C_i$ channels with different operators; further details of the block are provided in Eq. (\ref{eq:meta})
\begin{equation}
    \begin{split}
    X'_i &= X_i + TokenMixer(X_i), \\
    X''_i &= X'_i + ChannelMixer(X'_i),
    \end{split}
\label{eq:meta}
\end{equation}
where $TokenMixer(.)$ operator is configured as a convolution mixer or self-attention (see Table \ref{tab:arch_variant}). $ChannelMixer(.)$ contains the Multi-Layer Perceptron (MLP) network  that is conducted by two linearly fully connected layers followed by Batch Normalization (BN) and a single activation function that can be expressed in Eq. (\ref{eq:ffn}) as follows:
\begin{equation}
    MLP(X'_i)=BN\Big(\sigma\big(BN(X'_i*W_e)\big)*W_r\Big),
    \label{eq:ffn}
\end{equation}
where $W_e\in\mathbb{R}^{(C_i)\times rC_i}$ and $W_r\in\mathbb{R}^{(rC_i)\times C_i}$ are the layer weights, $r$ is the expansion ratio of the fully connected layer with a default value of 3. Operation $\sigma$ is chosen using the activation function $GELU(.)$.
\subsection{Structural Reparameterization}
\subsubsection{Fused Batch Normalization}
In CNN-based facial recognition systems, convolutional layers are commonly combined with Batch Normalization (BN) layers \cite{chen2018mobilefacenets, martindez2019shufflefacenet}. Adding BN after convolution is fundamental for improving convergence and reducing overfitting in training. However, it also elevates complexity and latency during inference. To resolve this, the BN is merged into the preceding convolution layer to form the FaceLiVT.
Convolutional layer with kernel size $K$, the weight matrix $W$ is defined as $W \in \mathbb{R}^{C_o \times C_i \times K \times K}$, and the bias $b$ as $b \in \mathbb{R}^D$, where $C_i$ and $C_o$ are the input and output channel dimensions, respectively. The convolution on feature $X \in \mathbb{R}^{N \times C_i \times H \times W}$ is followed by BN, involving the accumulated mean $\mu$, accumulated standard deviation $\sigma$, feature scale $\gamma$, bias $\beta$, and convolution operation $*$, as described in Eq.(\ref{eq:bnconv}).
\begin{equation}
    BN(Conv(X))=\gamma\frac{(W*X+b)-\mu}{\sigma}+\beta.
    \label{eq:bnconv}
\end{equation}
As convolutions followed by BN during inference are linear operations, these can be merged into a single convolution layer with integrated BN, represented by Eq. (\ref{eq:fusedbn}): 
\begin{equation}
    BNConv(x)=W' \cdot X + b',
    \label{eq:fusedbn}
\end{equation}
where the transformed weight is $W' = W \frac{\gamma}{\sigma}$ and the adjusted bias is $b' = (b - \mu) \frac{\gamma}{\sigma} + \beta$. BN is merged into the preceding convolutional layer across all branches, leaving only convolution in the architecture.  
\subsubsection{Reparameterized Token Mixer (RepMix)}
The concept of convolutional mixing was initially presented in ConvMixer\cite{trockman2022patches}. For an input tensor $X_i$, the mixing block within the layer was formulated as
\vspace{-0.2\baselineskip}
\begin{equation}
    X'_i = X_i + BN(\sigma(DWC(X_i))),
\end{equation}
where \( \sigma \) denotes a non-linear activation function, and DWC is a DepthWise Convolutional layer. Although this configuration is proved quite effective, the authors of \cite{vasu2023fastvit} modified the sequence of operations and omitted the non-linear activation. To improve the RepMix, we enhance it with a $1\times1$ DWC after $k\times k$ DWC, which can enhance learnability during training.
\vspace{-0.2\baselineskip}
\begin{equation}
    X'_i = X_i + \left\{BN(DWC_{k\times k}(X_i) + DWC_{1\times 1}(X_i))\right\}.
\end{equation}
To reduce the computational load and memory requirements of both the skip connection and the $1\times1$ DWC, these can be reparameterized into a single depthwise convolutional layer at inference time, which is especially beneficial for mobile devices.
\subsection{Multi Head Self Attention}
In vision transformers, the Multi-Head Self-Attention (MHSA) mechanism allows the model to evaluate token significance in a sequence for prediction and context. For an input sequence $X$ with $N$ tokens, MHSA computes key $K$, query $Q$, and value $V$ through linear transformations, with $K, Q, V \in\mathbb{R}^{B\times H_e \times N \times C}$, where $B$ is the batch size, $H_e$ is the number of heads, $N$ is tokens, and $C$ is the channel dimension. Details of MHSA are outlined in Eq. (\ref{eq:mhsa}).
\vspace{-0.2\baselineskip}
\begin{equation}
    MHSA(Q,K,V)=Concat(SA_0,..,SA_{H_e})W_o.
    \label{eq:mhsa}
\end{equation}
\begin{equation}
    SA=Softmax\bigg(\frac{QK^T}{\sqrt{C}}\bigg)V.
    \label{eq:sa}
\end{equation}
where $SA$ refers to the self-attention operation in each head.  It calculates a weighted average of the values based on a similarity score between token pairs  as described in Eq.\ref{eq:sa}.

\subsection{Multi Head Linear Attention}
To reduce the computational demands while maintaining the understanding of long-range context, we present "Multi-Head Linear Attention" (MHLA). Let $X \in \mathbb{R}^{B\times C \times H \times W}$ denote the feature map with a resolution of $H \times W$ and $C$ channels. It is first transformed to a 1D representation with $N$ tokens, making it $X \in \mathbb{R}^{B\times C \times N}$. Subsequently, it will be divided across channels into $H_e$ heads, resulting in $X_{H_e} \in \mathbb{R}^{B\times \frac{C}{H_e} \times N}$. The details of MHLA are defined as follows:
\vspace{-0.2\baselineskip}
\begin{equation}
    MHLA(Q,K,V)=Concat(LA_0,..,LA_{H_e}),
    \label{eq:mhla}
\end{equation}
where $LA_{H_e}$ in each head comprises a sequence of two weighted linear operations with non-linear activation functions to evaluate spatial relationships among input tokens. $LA_{H_e}$ in each head can be expressed as:
\vspace{-0.2\baselineskip}
\begin{equation}
    LA_{H_e}(X_{H_e})=\Big(W_o\big(\sigma(X_{He}\cdot W_i)\big)\Big),
    \label{eq:la}
\end{equation}
with $W_i \in \mathbb{R}^{N\times Nr}$ and $W_o \in \mathbb{R}^{Nr\times N}$ denoting the linear weights. In addition, $Nr$ is the number of tokens with the expansion ratio $r$. When $X_{He}$ is multiplied by weights $W_i$ and $W_o$, the computational complexity, which depends on $Nr$ with total complexity of $\Omega(MHLA)=2(NNr)C$. The total complexity of MHLA is lower than the $\Omega(MHSA)=4NC^2+2N^2C$ \cite{liu2021swin}.

\section{Experiments}
\label{sec:result}
\subsection{Traning and Testing Details}
FaceLiVT trained with the Glint360K dataset \cite{an2021partial}, which consists of pre-aligned $112\times112$ resolution facial images. They were transformed into tensors and normalized between -1 and 1. Training was performed in 20 and 40 epoch with a batch size of 256 in each of three Nvidia RTX A6000 (40GB) GPUs in a distributed setting. AdamW optimizer with a learning rate of $6\times10^{-3}$, the CosFace \cite{wang2018cosface} loss function, and a polynomial decay learning rate schedule were used with a 512-dimensional embedding size in the PartialFC  \cite{an2021partial} training algorithm.

We evaluated the proposed FaceLiVT model utilizing seven diverse benchmark datasets, including LFW \cite{huang2008labeled}, CFP-FP \cite{sengupta2016frontal}, AgeDB-30 \cite{moschoglou2017agedb}, IJB-B \cite{whitelam2017iarpa}, and IJB-C \cite{maze2018iarpa}. We provide the True Accept Rate (TAR) at a False Accept Rate (FAR) of 1e-4 for IJB-B and IJB-C datasets. In the inference speed test, the model was converted with coremltools and measured the latency on the iPhone 15 Pro.

\begin{table}[ht]
\begin{center}
\caption{All Variant FaceLiVT Model configurations. \#Blocks denotes the number of FaceLiVT blocks.}
\begin{tabular}{cccccccc}
\hline
\multicolumn{1}{c|}{\multirow{2}{*}{Stage}} & \multicolumn{1}{c|}{\multirow{2}{*}{Size}} & \multicolumn{1}{c|}{\multirow{2}{*}{Layer}} & \multicolumn{4}{c}{FaceLiVT}  \\ 
\cline{4-7} 
\multicolumn{1}{l|}{} & \multicolumn{1}{l|}{} & \multicolumn{1}{l|}{} & \multicolumn{1}{c|}{S} & \multicolumn{1}{c|}{M} & \multicolumn{1}{c|}{S-(Li)} & \multicolumn{1}{c}{M-(Li)} \\ 
\hline
\multicolumn{1}{c|}{\multirow{2}{*}{Stem}} & \multicolumn{1}{c|}{\multirow{2}{*}{$112^2$}} & \multicolumn{1}{c|}{\multirow{2}{*}{\begin{tabular}[c]{@{}c@{}}Conv\\ Dims($C_i$)\end{tabular}}} & \multicolumn{4}{c}{$[3 \times 3$, Stride $2]\times2$} \\ 
\cline{4-7}  
\multicolumn{1}{l|}{} & \multicolumn{1}{l|}{} & \multicolumn{1}{c|}{} & \multicolumn{1}{c|}{40} & \multicolumn{1}{c|}{64}  & \multicolumn{1}{c|}{40} & \multicolumn{1}{c}{64}  \\ 
\hline
\multicolumn{1}{c|}{\multirow{2}{*}{1}} & \multicolumn{1}{c|}{\multirow{2}{*}{$28^2$}} & \multicolumn{1}{l|}{Mixer}  & \multicolumn{4}{c}{RepMix $3 \times 3$ }  \\
\cline{3-7}
\multicolumn{1}{l|}{} & \multicolumn{1}{l|}{} & \multicolumn{1}{l|}{\#Blocks} & \multicolumn{1}{c|}{2}  & \multicolumn{1}{c|}{2}  & \multicolumn{1}{c|}{2} & \multicolumn{1}{c}{2} \\ 
\hline
\multicolumn{1}{c|}{\multirow{4}{*}{2}} & \multicolumn{1}{c|}{\multirow{4}{*}{$14^2$}} & \multicolumn{1}{l|}{Downspl} & \multicolumn{4}{c}{RepMix $3 \times 3$, Stride 2} \\ 
\cline{4-7} 
\multicolumn{1}{l|}{} & \multicolumn{1}{l|}{} & \multicolumn{1}{l|}{Dim($C_i$)} & \multicolumn{1}{c|}{80} & \multicolumn{1}{c|}{128}  & \multicolumn{1}{c|}{80} & \multicolumn{1}{c}{128}  \\ 
\cline{3-7} 
\multicolumn{1}{l|}{} & \multicolumn{1}{l|}{} & \multicolumn{1}{l|}{Mixer}  & \multicolumn{4}{c}{RepMix $3 \times 3$ }  \\
\cline{3-7} 
\multicolumn{1}{l|}{} & \multicolumn{1}{l|}{} & \multicolumn{1}{l|}{\#Blocks} & \multicolumn{1}{c|}{4}  & \multicolumn{1}{c|}{4}  & \multicolumn{1}{c|}{4} & \multicolumn{1}{c}{4} \\ 
\hline
\multicolumn{1}{c|}{\multirow{4}{*}{3}} & \multicolumn{1}{c|}{\multirow{4}{*}{$7^2$}} & \multicolumn{1}{l|}{Downspl} & \multicolumn{4}{c}{RepMix $3 \times 3$, Stride 2} \\ 
\cline{4-7} 
\multicolumn{1}{l|}{} & \multicolumn{1}{l|}{} & \multicolumn{1}{l|}{Dim ($C_i$)} & \multicolumn{1}{c|}{160} & \multicolumn{1}{c|}{256}  & \multicolumn{1}{c|}{160} & \multicolumn{1}{c}{256}  \\ 
\cline{3-7} 
\multicolumn{1}{l|}{} & \multicolumn{1}{l|}{} & \multicolumn{1}{l|}{Mixer}  & \multicolumn{2}{c|}{MHSA} & \multicolumn{2}{c}{MHLA}   \\
\cline{3-7}
\multicolumn{1}{l|}{} & \multicolumn{1}{l|}{} & \multicolumn{1}{l|}{\#Blocks} & \multicolumn{1}{c|}{6}  & \multicolumn{1}{c|}{6}  & \multicolumn{1}{c|}{6} & \multicolumn{1}{c}{6} \\ 
\hline
\multicolumn{1}{c|}{\multirow{4}{*}{4}} & \multicolumn{1}{c|}{\multirow{4}{*}{$4^2$}} & \multicolumn{1}{l|}{Downspl} & \multicolumn{4}{c}{RepMix $3 \times 3$, Stride 2} \\ 
\cline{4-7} 
\multicolumn{1}{l|}{} & \multicolumn{1}{l|}{} & \multicolumn{1}{l|}{Dim ($C_i$)} & \multicolumn{1}{c|}{320} & \multicolumn{1}{c|}{512}  & \multicolumn{1}{c|}{320} & \multicolumn{1}{c}{512}  \\ 
\cline{3-7} 
\multicolumn{1}{l|}{} & \multicolumn{1}{l|}{} & \multicolumn{1}{l|}{Mixer}  &  \multicolumn{2}{c|}{MHSA} & \multicolumn{2}{c}{MHLA} \\
\cline{3-7} 
\multicolumn{1}{l|}{} & \multicolumn{1}{l|}{} & \multicolumn{1}{l|}{\#Blocks} & \multicolumn{1}{c|}{2}  & \multicolumn{1}{c|}{2}  & \multicolumn{1}{c|}{2} & \multicolumn{1}{c}{2} \\ 
\hline
\multicolumn{3}{c|}{Classifier Head} & \multicolumn{4}{c}{Avg Pool, FC (512) }\\ \hline
\end{tabular}
\label{tab:arch_variant}
\end{center}
\end{table}

\subsection{Benchmarking Result}
Table \ref{tab:benchmark} provides a comparison of several face recognition models, including the FaceLiVT variants, against state-of-the-art models, with respect to parameters, computational cost (FLOP), accuracy on benchmark datasets, and latency measured on an iPhone 15 Pro. We categorized models based on the number of FLOP around 300-1100 M FLOP and $<$300 M FLOP. The FaceLiVT models, which utilize a Hybrid Vision Transformer (ViT) structure, demonstrate a balance between computational efficiency and accuracy performance. In comparison with conventional CNN-based models such as MobileFaceNet and ShuffleFaceNet, the FaceLiVT models typically achieve superior accuracy on rigorous benchmarks like CFP-FP, AgeDB-30, IJB-B, and IJB-C, highlighting the advantages of the Hybrid ViT architecture in managing diverse and complex facial variations.

Among the CNN-based baselines, MobileFaceNet and ShuffleFaceNet offer lightweight solutions with reasonable performance. However, their recognition accuracy decreases noticeably on more challenging datasets such as CFP-FP, AgeDB-30, and IJB-C. GhostFaceNet-V2 improves upon this with better generalization and a low latency of 0.71 ms. SwiftFaceFormer-L1, as a hybrid model, provides improved accuracy but still exhibits higher latency (1.50 ms), limiting its suitability for real-time applications.

In contrast, FaceLiVT variants achieve strong accuracy-latency trade-offs, especially those incorporating MHLA. For instance, FaceLiVT-M(LA) delivers high performance on all benchmark datasets with a similar inference speed from FaceLiVT-S that used MHSA. Moreover, it can achieve competitive accuracy with 8.6 $\times$ faster than EdgeFace-XS(0.6), the recent hybrid ViT for face recognition, and 21.2 $\times$ faster than the pure ViT-Based model. This underscores the capability of MHLA in enhancing computational efficacy while maintaining a satisfactory level of accuracy. It also indicates that MHLA can greatly improve real-time usability on mobile platforms. Besides its efficiency, MHLA may limit the model’s capacity to capture complex long-range dependencies compared to MHSA, especially in highly unconstrained environments that lead to slight performance degradation.

\begin{table*}[!ht]
\centering
\caption{Comparison of FaceLiVT Variant with State-Of-The-Art on Face Recognition Benchmark Dataset.}
\begin{tabular}{ m{3.7cm}|>\centering m{0.95cm}|>\centering m{0.8cm}|>\centering m{0.9cm}|>\centering m{0.8cm}|>\centering m{0.9cm}|>\centering m{0.8cm}|>\centering m{1.2cm}|>\centering m{0.8cm}|>\centering m{0.8cm}|c } 
\hline
\multirow{2}{*}{Model} & \multirow{2}{*}{Type} & Param & FLOP & Train & \multirow{2}{*}{LFW} & CFP & Age &\multicolumn{2}{c|}{IJB}& Lat \\ \cline{9-10}
& & (M) & (M) & Epoch & & -FP &DB-30& B & C &(ms) \\ \hline
ViT-S \cite{dan2023transface} & ViT & 86.6 & 5,713 & 40 & 99.8 & 98.9 & 98.3 & - & 96.7 & 14.23 \\
TransFace-S \cite{dan2023transface}  & ViT & 86.7 & 5,824 &  40 & 99.9 & 98.9 & 98.5 & - & 97.3 & 14.31 \\
MobileFaceNet\cite{martinez2021benchmarking}  & CNN & 2.0 & 933 & 20 & 99.7 & 96.9 & 97.6 & 92.8 & 94.7 & 0.77\\
MobileFaceNetV1\cite{martinez2021benchmarking} & CNN & 3.4 & 1100 & 20 & 99.4 & 95.8 & 96.4 & 92.0 & 93.9 & 0.81 \\
SwiftFaceFormer-L1 \cite{luevano2024swiftfaceformer} & Hybrid & 11.8 & 805 & 35 & 99.7 & 96.7 & 97.0 & 91.8 & 93.8 & 1.50 \\
ShuffleFaceNet-1.5\cite{martinez2021benchmarking} & CNN & 2.6 & 577 & 20 & 99.7 & 96.9 & 97.3 & 92.3 & 94.3 & 0.69\\ 
EdgeFace-S(0.5) \cite{george2024edgeface} & Hybrid & 3.6 & 306 & 50 & 99.8 & 95.8 & 96.9 & 93.6 & 95.6 & 10.21\\ 
GhoseFaceNet-V2 \cite{alansari2023ghostfacenets} & CNN & 6.88 & 272 & 50 & 99.9 & 98.9 & 98.5 & 95.7 & 97.0 & 0.71 \\
\rowcolor{gray!20}
FaceLiVT-M         & Hybrid & 14.3 & 569 & 20 & 99.8 & 97.1 & 97.2 & 93.4 & 95.0 & 1.11\\
\rowcolor{gray!20}
FaceLiVT-M-(LA)    & Hybrid & 9.75 & 386 & 20/ 40 & 99.7/ 99.8 & 96.0/ 97.2 & 96.7/ 97.6 & 92.5/ 93.7 & 94.1/ 95.7 & 0.67 \\
\hline
ShuffleFaceNet-0.5 \cite{martinez2021benchmarking} & CNN & 1.4 & 66.9 & 20 & 99.2 & 92.6 & 93.2 & - & - & 0.45\\
EdgeFace-XS(0.6) \cite{george2024edgeface} & Hybrid & 1.77 & 154 & 50 & 99.7 & 94.4 & 96.0 & 92.7 & 94.8 & 5.82 \\ 
\rowcolor{gray!20}
FaceLiVT-S       & Hybrid & 5.89 & 237 & 20 & 99.7 & 95.2 & 96.3 & 89.1 & 89.7 & 0.61 \\
\rowcolor{gray!20}
FaceLiVT-S-(LA) & Hybrid & 5.05 & 160 & 20 / 40 & 99.6 / 99.7 & 94.6/ 95.1 & 95.6 / 96.6 & 83.4 / 91.2 & 82.5 / 92.7 & 0.47 \\
\hline
\end{tabular}
\label{tab:benchmark}
\end{table*}

\subsection{Ablation Study}
We conduct a 20 epoch ablation study to identify two key factors affecting the performance of FaceLiVT-S-(LA): structural reparameterization and the count of heads ($H_e$) MHLA mechanism. According to Table \ref{tab:abl-1}, structural reparameterization is pivotal in enhancing the latency of FaceLiVT-S-(LA). Eliminating reparameterization for residual and BN raises the latency from 0.47 ms to 0.50 ms and 0.60 ms, thus highlighting the effectiveness of this approach in boosting computational speed. Additionally, removing $DWC_{1\times1}$ as weight refinement emphasizes its significance in accuracy around 0.9\% in CFP-PP and 1.0\% for Age DB-30. Importantly, these enhancements do not affect the model parameters (Param) or computational cost (FLOP), indicating that these methods refine the processing pipeline without impacting the model's overall complexity.
\begin{table}[!ht]
\centering
\caption{Ablation of RepMixer block in FaceLiVT-S-(LA)}
\footnotesize
\begin{tabular}{ m{1.8cm}|c|c|c|c|c|c } 
\hline
\multirow{2}{*}{Ablation} & Par & FLOP & \multirow{2}{*}{LFW} & CFP & Age & Lat \\ & (M) & (M) &  & -FP & DB-30 & (ms) \\ \hline
\rowcolor{gray!20}
Baseline & 5.05 & 160 & 99.6 & 94.9 & 95.6 & 0.47 \\
\hline
w/o Res Rep & 5.05 & 160 & 99.6 & 94.9 & 95.6 & 0.50 \\
w/o fused BN & 5.10 & 160 & 99.6 & 94.9 & 95.6 & 0.60 \\
w/o $DWC_{1\times1}$  & 5.05 & 160 & 99.6 & 93.5 & 94.6 & 0.47 \\
\hline
\end{tabular}
\label{tab:abl-1}
\end{table}
Table \ref{tab:abl-2} shows  the impact of $H_e$, the number of heads in MHLA, is assessed. Reducing the $H_e$ to 8 decreases the parameter count to 4.09 M and slightly reduces FLOPs to 157 M, which enhances latency to 0.41 ms. Nevertheless, this adjustment leads to a minor decline in accuracy, with model performance registering at 99.6\% on LFW and 93.9\% on CFP-FP. Conversely, increasing $H_e$ to 16 enhances accuracy to 94.6\% on CFP-FP and 95.6\% on AgeDB-30, accompanied by a slight rise in computational demands and latency (now 0.48 ms). This indicates that augmenting $H_e$ bolsters the model's capability to grasp complex features at the expense of a slight decrease in runtime efficiency.

The ablation study indicates that structural reparameterization methods and the selection of $H_e$ in MHLA are critical factors for balancing accuracy and latency in FaceLiVT-S-(LA). The best configuration, which includes fused BN, residual reparameterization, and $H_e=16$, provides an advantageous trade-off by achieving high accuracy with minimal latency, rendering it suitable for real-time applications.

\begin{table}[!ht]
\centering
\caption{Ablation of FaceLiVT-S(LA) that using MHLA, The ablation shows the effect of number head $H_e$}
\begin{tabular}{ c|c|c|c|c|c|c } 
\hline
\multirow{2}{*}{$H_e$} & Par & FLOP & \multirow{2}{*}{LFW} & CFP & Age & Lat \\ & (M) & (M) &  & -FP &DB-30& (ms) \\ \hline
8  & 4.09 & 157 & 99.6 & 93.9  & 95.0 & 0.41 \\
\rowcolor{gray!20}
16 & 5.05 & 160 & 99.6  & 94.6  & 95.6 & 0.48 \\
\hline
\end{tabular}
\label{tab:abl-2}
\end{table}

\section{Conclusion}
\label{sec:conclusion}

The paper introduces FaceLiVT, a CNN-Transformer architecture with structural reparameterization and Multi-Head Linear Attention (MHLA) for effective face recognition on mobile platforms. Experiments on benchmarks such as LFW, AgeDB-30, CFP-FP, IJB-B, and IJB-C revealed the superior accuracy-latency balance over other lightweight models. MHLA significantly boosts inference speed while maintaining competitive performance, and reparameterization reduces computational cost without compromising accuracy. Although MHLA enhances speed and efficiency, its ability in complex long-range dependencies is less robust than full self-attention, affecting performance in environments with occlusions. Future work could explore MHLA to retain efficiency while enhancing contextual understanding. 

\bibliographystyle{IEEEbib}
\bibliography{refs}

\end{document}